\documentclass[letterpaper]{article} 
\usepackage{aaai2026}  
\usepackage{times}  
\usepackage{helvet}  
\usepackage{courier}  
\usepackage[hyphens]{url}  
\usepackage{graphicx} 
\usepackage{natbib}  
\usepackage{caption} 
\usepackage{algorithm}
\usepackage{algorithmic}

\usepackage{amssymb}

\usepackage{tabularray}
\usepackage{booktabs}
\usepackage{multirow}
\usepackage{amsmath}

\usepackage{newfloat}
\usepackage{listings}
\DeclareCaptionStyle{ruled}{labelfont=normalfont,labelsep=colon,strut=off} 
\floatstyle{ruled}
\newfloat{listing}{tb}{lst}{}
\floatname{listing}{Listing}
\title{P2S: Probabilistic Process Supervision for General-Domain Reasoning Question Answering}
\author{
    \textbf{Wenlin Zhong\textsuperscript{\rm 1}, Chengyuan Liu\textsuperscript{\rm 2}, Yiquan Wu\textsuperscript{\rm 3}\thanks{Corresponding author.}, Bovin Tan\textsuperscript{\rm 3}, Changlong Sun\textsuperscript{\rm 3}, Yi Wang\textsuperscript{\rm 4}, Xiaozhong Liu\textsuperscript{\rm 5}, Kun Kuang\textsuperscript{\rm 2}}
}

\affiliations{\fontsize{10}{12}\selectfont 

\textsuperscript{1}School of Software Technology, Zhejiang Unirersity\\
\textsuperscript{2}College of Computer Science and Technology, Zhejiang Unirersity \\
\textsuperscript{3}Guanghua Law School, Zhejiang University\\
\textsuperscript{4}Chongqing Ant Consumer Finance Co,. Ltd , Ant Group \\
\textsuperscript{5}Worcester Polytechnic Institute, Worcester, USA\\

\fontsize{10}{12}\selectfont \{22451152, liucy1, wuyiquan, bovintan, 11921173, kunkuang\}@zju.edu.cn,
\fontsize{10}{12}\selectfont haonan.wy@myxiaojin.cn, xliu14@wpi.edu
}

\usepackage{bibentry}

\begin{document}

\maketitle

\begin{abstract}
While reinforcement learning with verifiable rewards (RLVR) has advanced LLM reasoning in structured domains like mathematics and programming, its application to general-domain reasoning tasks remains challenging due to the absence of verifiable reward signals. To this end, methods like Reinforcement Learning with Reference Probability Reward (RLPR) have emerged, leveraging the probability of generating the final answer as a reward signal. However, these outcome-focused approaches neglect crucial step-by-step supervision of the reasoning process itself. To address this gap, we introduce Probabilistic Process Supervision (P2S), a novel self-supervision framework that provides fine-grained process rewards without requiring a separate reward model or human-annotated reasoning steps. During reinforcement learning, P2S synthesizes and filters a high-quality reference reasoning chain (gold-CoT). The core of our method is to calculate a Path Faithfulness Reward (PFR) for each reasoning step, which is derived from the conditional probability of generating the gold-CoT's suffix, given the model's current reasoning prefix. Crucially, this PFR can be flexibly integrated with any outcome-based reward, directly tackling the reward sparsity problem by providing dense guidance. Extensive experiments on reading comprehension and medical Question Answering benchmarks show that P2S significantly outperforms strong baselines.
\end{abstract}


\section{Introduction}

Large-scale Reinforcement Learning with Verifiable Rewards (RLVR) has emerged as a promising paradigm to advance the reasoning capabilities of Large Language Models (LLMs)~\cite{guo2025deepseek, Luo_et_al_2025a, xu2025copyright}. This approach has fueled a major leap forward, particularly in structured, verifiable domains such as mathematics and programming~\cite{shao2024deepseekmath, havrilla2024teaching, kumar2024training, cao2024survey}. Within this paradigm, LLMs are trained using verifiable rewards computed directly from the model's own final outcomes, such as matching ground truth answers, passing unit tests, or selecting the correct option in multiple-choice questions (MCQ)~\cite{Schulman_et_al_2017, setlur2024rewarding, xie2025logic}.

\begin{figure}
    \centering
    \includegraphics[width=1\linewidth]{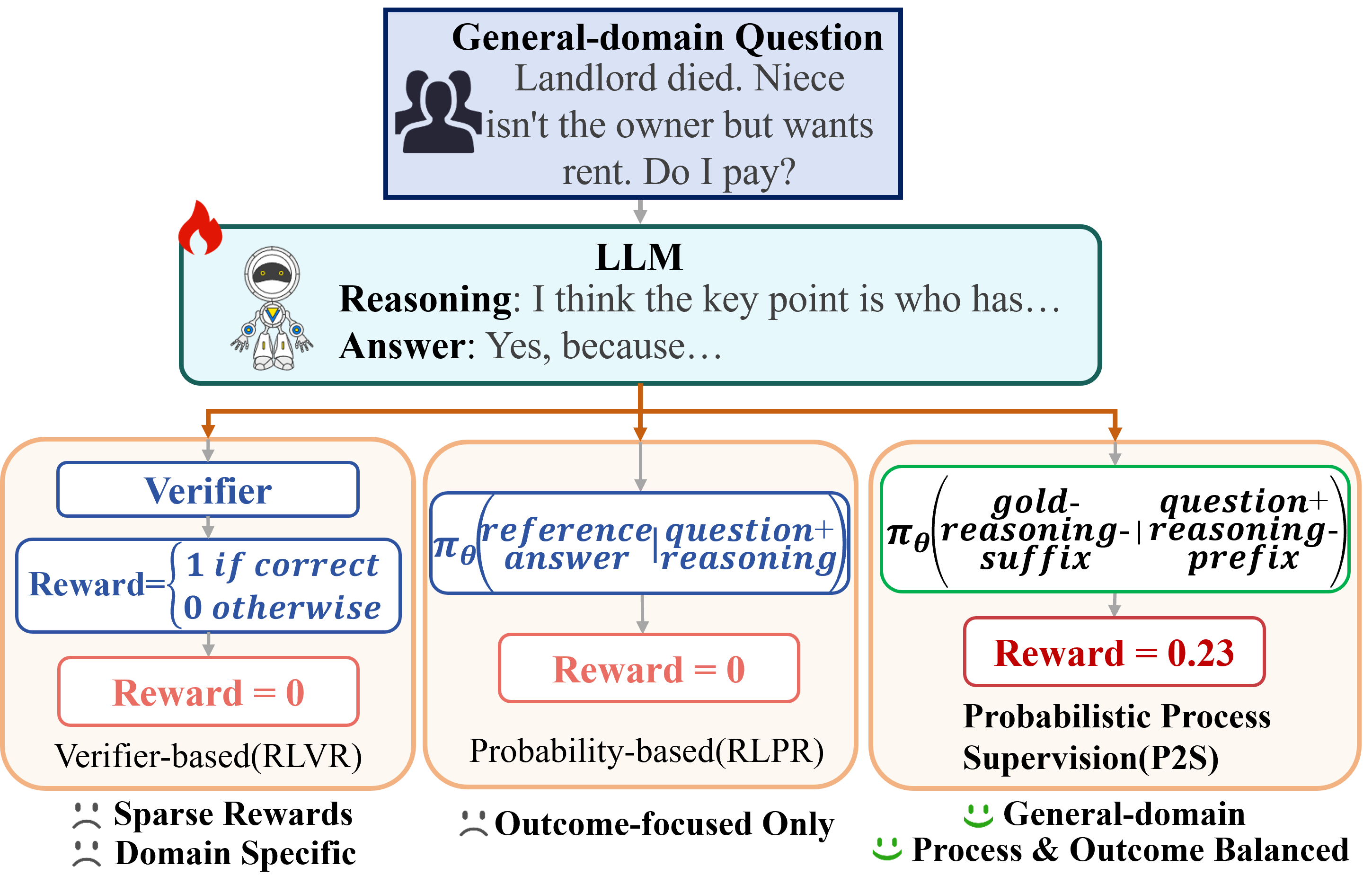}
    \caption{Comparing reward mechanisms: P2S rewards the entire reasoning process.}
    \label{fig:fig1}
\end{figure}

While RLVR has excelled in specific domains, its success does not readily transfer to general-domain reasoning. The free-form and stylistically diverse nature of answers in these tasks makes designing a direct, verifiable reward signal a challenge. Conventional solutions are inadequate: manually engineering reward functions is unscalable~\cite{zeng2025simplerl,hu2025open}, and training a specialized LLM as a verifier~\cite{ma2025general} demands extensive data annotation, yields unsatisfactory reward quality, and complicates the training pipeline. A more promising direction, Reinforcement Learning with Reference Probability Reward (RLPR)~\cite{xu2025direct,yu2025rlpr,zhou2025reinforcing}, leverages the generation probability of the final answer as a reward. However, all these outcome-focused methods share critical flaws: they neglect step-by-step process supervision, which can lead models to discover ``shortcut'' solutions via flawed logic and exacerbates reward sparsity in complex problems.

As shown in Figure~\ref{fig:fig1}, we compare P2S with RLVR and RLPR in general-domain QA. In contrast to domain-specific, sparse-reward verifiers (Figure 1, left) and purely outcome-focused RLPR (Figure 1, center), we argue that the supervisory signal within the reasoning chain itself remains a valuable, untapped resource. Therefore, we aim to design a new reward mechanism that moves beyond sparse outcomes and learns directly from the step-by-step reasoning process, providing more effective and fine-grained supervision for general-domain tasks.

To remedy this oversight, directly supervising the reasoning process is a natural next step. However, prevailing approaches introduce significant burdens. Training a separate reward model necessitates a large corpus of human-annotated or LLM-generated preference data ~\cite{lightman2023let}, incurring substantial annotation and computational costs. Alternatively, Monte Carlo search-based ~\cite{wang2023math} methods, which score each step via multiple rollouts to a terminal state, face severe scalability challenges. The required sample count grows prohibitively with the reasoning chain's length, leading to immense computational overhead. \textbf{This highlights a crucial need for a process supervision method that is both low-cost and computationally tractable.}
 
Our work addresses this challenge by introducing Probabilistic Process Supervision (P2S), a low-cost, self-bootstrapping mechanism that provides fine-grained, process-level supervision by scoring and learning from the model's own reasoning paths, eliminating the need for external reward models or human annotations. To achieve this, we introduce two core techniques.

First, we introduce a dynamic gold-CoT synthesis mechanism. For each problem, we prompt the model with the question and its ground truth answer to generate multiple candidate reasoning paths. These paths are then filtered based on both their final answer's correctness and their internal reasoning quality, creating a high-quality, dynamically updated set of reference chains that adapts to the model's evolving capabilities. Second, we introduce the Path Faithfulness Reward (PFR), our core innovation for dense, step-level supervision. PFR measures how ``faithful'' a generated reasoning path is to a reference gold-CoT. At each step of the generated path, PFR calculates the conditional probability of completing the rest of the gold-CoT from that point. This step-wise score quantifies whether the model is on a logically sound trajectory. These scores are then aggregated into a sample-level reward that penalizes early deviations and rewards consistent logical progression, thereby directly providing the dense, process-level signal needed to overcome reward sparsity. Finally, P2S operates within a flexible reinforcement learning paradigm. Our process-based PFR can be seamlessly combined with any outcome-based reward, creating a hybrid signal. This joint optimization ensures the model learns not only from successful outcomes but also from the quality of its reasoning process, providing a dense and robust reward signal even when all samples in a batch are incorrect.

Extensive experiments on diverse benchmarks, including general-domain reading comprehension and medical QA, demonstrate that P2S significantly outperforms strong baselines. Our main contributions are summarized as follows:

\begin{itemize}
    \item We explore the challenging task of reinforcement learning for reasoning in general-domain QA, where traditional verifiable rewards are often unavailable. we identify the limitations of current outcome-focused approaches and propose a new direction centered on process-level supervision derived from the model's own generation probabilities.

    \item We introduce Probabilistic Process Supervision (P2S), a novel self-supervision framework that generates fine-grained, process-level rewards without costly external reward models or human annotations. At its core, P2S leverages two innovations: a dynamic Gold-CoT synthesis mechanism and our Path Faithfulness Reward (PFR).

    \item We demonstrate through extensive experiments on diverse benchmarks, including general-domain reading comprehension and medical QA, that P2S consistently and significantly outperforms strong state-of-the-art baselines.
\end{itemize}

\section{Related Work}
\subsection{Reinforcement Learning for Reasoning}
To advance beyond simple prompting for Chain-of-Thought (CoT) reasoning~\cite{kojima2022large, wei2022chain}, recent paradigms directly train LLMs, notably via reinforcement learning (RL) on reasoning traces~\cite{shao2024deepseekmath, he2025skywork}. A successful branch, RLVR, excels in structured domains like math and code by using deterministic, binary outcome rewards from verifiers ~\cite{guo2025deepseek, yu2025dapo, ye2025limo}. However, this reliance on verifiers makes RLVR unsuitable for general-domain reasoning, where such clear verification is often impossible.
\subsection{Reasoning in General Domains}
To enable reinforcement learning in general reasoning domains without clear verifiers, research has focused on designing reliable reward signals. One major direction is to train an external generative reward model to act as a judge \cite{mahan2024generative,ma2025general}, which introduces the overhead of developing and maintaining an additional reward model during RL training. A competing approach avoids this by using the policy model's internal feedback as a reward, leveraging signals such as self-certainty or the probability of the ground truth answer as a reward signal. \cite{xu2025direct,yu2025rlpr,zhou2025reinforcing}.

\subsection{Process Reward Supervision}

Process supervision improves LLM reasoning consistency by rewarding intermediate steps. While training reward models on human-annotated steps~\cite{li2024pspo, lightman2023let} is costly and unscalable, search-based alternatives like Monte Carlo search estimate step values via rollouts~\cite{wang2023math,guo2025segment}. However, these methods incur prohibitive computational costs that scale poorly with reasoning length.

\section{Preliminaries}

We first introduce the reasoning optimization with RL, upon which many works build to perform RLVR. Then, we introduce the emerging approach of RLPR~\cite{xu2025direct,yu2025rlpr,zhou2025reinforcing}.


%
\subsection{Reasoning Optimization With RL}

In order to enhance the reasoning ability of large models, we adopt Group Relative Policy Optimization (GRPO)~\cite{shao2024deepseekmath} following the recent advancements such as DeepSeek-R1~\cite{guo2025deepseek}. Given a question-answer pair $(q, a)$, a behavior policy $\pi_{\theta_{\text{old}}}$ samples a group of $G$ individual responses $\{o_i\}_{i=1}^G$. The GRPO objective updates model parameters $\theta$ as follows:

\begin{equation}  \label{eq:grpo}
\begin{split}
    \mathcal{J}_{\text{GRPO}}(\theta) =& \mathbb{E}_{\substack{(q,a)\sim D, \{o_i\}_{i=1}^G \sim \pi_{\theta_{\text{old}}}(\cdot|q)}} 
    \\ 
    & \Bigg[ \frac{1}{G} \sum_{i=1}^G \frac{1}{|o_i|} \sum_{t=1}^{|o_i|} \bigg\{ \min \bigg[ \frac{\pi_\theta(o_{i,t}|q, o_{i,<t})}{\pi_{\theta_{\text{old}}}(o_{i,t}|q, o_{i,<t})} \hat{A}_{i,t}, \\
    & \text{clip}\left(\frac{\pi_\theta(o_{i,t}|q, o_{i,<t})}{\pi_{\theta_{\text{old}}}(o_{i,t}|q, o_{i,<t})}, 1-\epsilon, 1+\epsilon\right) \hat{A}_{i,t} \bigg] \\
    & - \beta \mathbb{D}_{\text{KL}}(\pi_\theta \| \pi_{\text{ref}})  \bigg\} \Bigg]
\end{split}
\end{equation}

The key distinction of GRPO is its advantage estimation for the $t$-th token in the $i$-th output, $\hat{A}_{i,t}$. This involves a structured comparison across a group of $G$ outputs $\{\mathbf{o}_i\}_{i=1}^G$ sampled for the same prompt. Given corresponding rewards $\{R_i\}_{i=1}^G$, the advantage is estimated as:
\begin{equation}
\label{eq:advantage}
\hat{A}_{i,t} = \frac{r_i - \text{mean}(\{R_i\}_{i=1}^G)}{\text{std}(\{R_i\}_{i=1}^G)}
\end{equation}
In the context of RLVR, the reward $r_i$ is typically a verifiable signal, such as 1 if the final answer is correct and 0 otherwise. 
This group-normalized formulation steers the policy to assign higher probabilities to trajectories that outperform their peers within the same generation batch.


\subsection{Reinforcement Learning with Reference Probability Reward (RLPR)}

\label{sec:rlpr}

To address the scalability limitations of RLVR, a recent trend in general-domain reasoning is to adopt reinforcement learning paradigms that use probability-based reward signals. It leverages the LLM's own knowledge.

In a typical RLPR setup, for a given input query $\mathbf{q}$, the policy model $\pi_\theta$ first generates a full response $\mathbf{o}$, which includes both a reasoning path $\mathbf{z}$ and a final answer $y$. The reward is not based on the correctness of the generated answer $y$. Instead, it is computed from the model's conditional probability of generating the tokens of the ground truth answer $y^*$, given the generated reasoning path $\mathbf{z}$. This can be formally expressed as the aggregated log-probability:
\begin{equation}
\label{eq:rlpr_reward}
r_{\text{RLPR}} = \sum_{t=1}^{|y^*|} \log \pi_{\theta}(y^*_t | \mathbf{q}, \mathbf{z}, y^*_{<t})
\end{equation}
where $y^*_t$ is the $t$-th token of the ground truth answer.

\begin{figure*}[t]  
    \centering
    \includegraphics[width=1.0\textwidth]{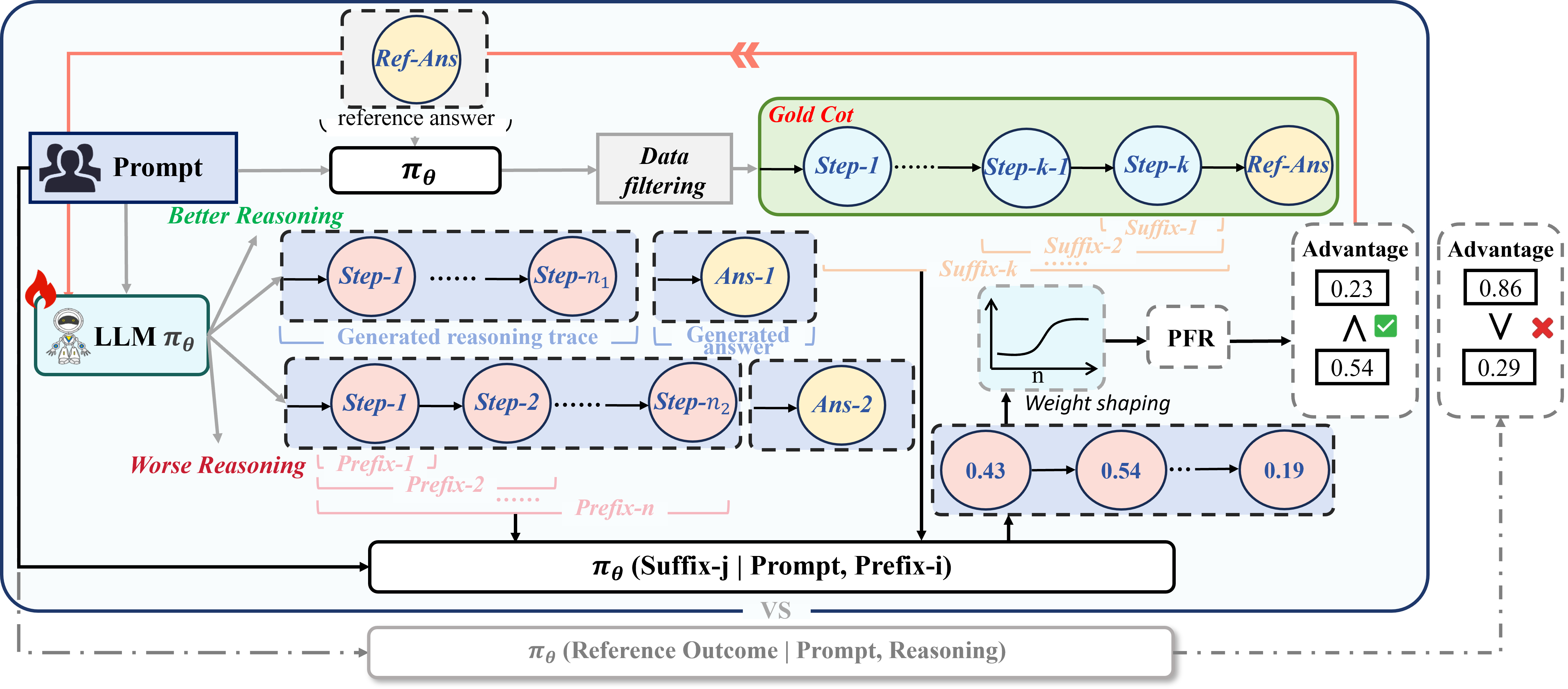} 
    \caption{An overview of our Probabilistic Process Supervision (P2S) framework. (1) Gold-CoT Synthesis (Top): A dynamic reference path (Gold-CoT) is created by generating and filtering the policy model's own reasoning outputs. (2) PFR Calculation (Bottom): For each new trace, a step-wise Path Faithfulness Reward (PFR) is computed by aligning it against the Gold-CoT. (3) Reward Shaping $\&$ Aggregation: The step-wise rewards are shaped using a sigmoid function to assign progressively higher weights to later reasoning steps. These weighted scores are then summed to produce the final, sample-level Path Faithfulness Reward (PFR) used for policy optimization.}
    \label{fig:main} 
\end{figure*} 

\section{Methodology}
\label{sec:method}

In this section, we begin by formally defining the problem, then outline the overall architecture of Probabilistic Process Supervision (P2S) framework, and finally, detail its core components.


\subsection{Problem Definition}
\label{sec:problem_definition}
We consider the task of learning a reasoning policy for general-domain question answering. Formally, we are given a dataset $\mathcal{D} = \{ (q_i, y^*_i) \}_{i=1}^{N}$, where $q_i$ is a question or prompt, and $y^*_i$ is its corresponding ground-truth final answer. A key characteristic of these tasks is their diversity, spanning multiple domains and featuring answers that are free-form text of varying lengths and styles.

Our goal is to learn a policy $\pi_{\theta}$ that, given a prompt $q$, generates a logically sound reasoning path $\mathbf{z} = (\mathbf{z_1}, \mathbf{z_2}, \dots, \mathbf{z_T},)$ which culminates in a final answer $y$. This diversity in the target answers $y^*$ makes exact string matching an unsuitable objective. Therefore, our ultimate goal is to maximize the semantic similarity between the generated answer $y$ and the ground-truth $y^*$.

\subsection{Overall Architecture}
\label{sec:overall_architecture}
As illustrated in Figure~\ref{fig:main}, our Probabilistic Process Supervision (P2S) framework operates as a self-improving loop that provides dense, process-level rewards for policy optimization. Firstly, within each iteration of the GRPO, a dynamic Gold-CoT synthesis mechanism leverages the current policy $\pi_{\theta}$, guided by a ground truth answer, to generate and filter multiple candidate reasoning paths. This yields a high-quality set of Gold-CoTs specifically tailored for the current learning state. Concurrently, for each generated reasoning trace in the batch, our Path Faithfulness Reward (PFR) is computed by aligning it against a reference Gold-CoT and calculating step-wise conditional probabilities. These step-wise rewards are then weighted and aggregated into a single, sample-level process reward and used to update the policy $\pi_{\theta}$, which provides a nuanced score for the entire reasoning path.

\subsection{Dynamic Gold-CoT Synthesis and Filtering}
\label{sec:gold_cot_generation}

To ensure a high-fidelity and adaptive supervision signal, P2S dynamically synthesizes and filters reference reasoning paths (Gold-CoTs) in each training iteration. This process involves two main steps: Candidate Synthesis and Quality-Based Filtering.

\paragraph{Candidate Synthesis.}To encourage the model to explore paths that lead to the correct answer, we prompt the policy model $\pi_\theta$ with both the query $q$ and the ground truth answer $y^*$ to generate a diverse set of $K$ candidate reasoning paths, $\{\mathbf{o}_k\}_{k=1}^K$ during this synthesis stage. This guided generation helps to efficiently sample trajectories within the vicinity of the correct solution space.

\begin{equation}
    \{\mathbf{o}_k\}_{k=1}^K \sim \pi_\theta(\cdot | q, y^*)
\end{equation}

\paragraph{Quality-Based Filtering.}
Simply generating paths guided by the ground truth answer $y^*$ is insufficient, as they may still be logically flawed, trivial, or fail to reach the correct final answer. Therefore, a filtering stage is crucial to isolate only the highest-quality candidates.

First, we discard any candidate $\mathbf{o}_k$ that does not adhere to a required structural format. Following the standard of~\cite{guo2025deepseek}, this format is $<$think$>$Reasoning$<$/think$>$$<$answer$>$Answer$<$/answer$>$. This preliminary step ensures that the reasoning path $\mathbf{z}_k$ and the final answer $y_k$ can be reliably parsed. Let the set of format-correct candidates be $\mathcal{C}_{\text{formatted}}$.

Then, for each candidate in $\mathcal{C}_{\text{formatted}}$, we compute a quality score $S_k$ as the conditional log-probability of generating the ground truth answer $y^*$ given the candidate's reasoning $\mathbf{z}_k$:
\begin{equation} \label{eq:filtering_score}
    S_k = \sum_{t=1}^{|y^*|} \log \pi_\theta(y^*_t | \mathbf{q}, \mathbf{z}_k, y^*_{<t})
\end{equation}
For each problem $q$, the definitive gold-CoT $\mathbf{o}^*$ is then selected by finding the candidate that maximizes this score:
\begin{equation*}
    \mathbf{o}^* = \underset{\mathbf{o}_k \in \mathcal{C}_{\text{formatted}}}{\arg\max} \, S_k
\end{equation*}

The resulting set of candidates $\mathcal{C}_{\text{gold}}$ forms a dynamic and high-quality benchmark for the current training step. This self-improving mechanism creates a virtuous cycle: as the policy model $\pi_\theta$ improves, so does the quality of its self-generated supervision.


\subsection{Path Faithfulness Reward (PFR)}

The core of our P2S framework is the Path Faithfulness Reward (PFR), which provides a dense, step-level reward to guide the model's reasoning process. 
The central intuition is that a high-quality reasoning prefix should significantly increase the likelihood of generating a subsequent, logically sound reasoning segment from a verified gold-CoT.

We first segment the generated chain $\mathbf{z}$ into a sequence of up to $\text{MAX\_STEP\_NUM}$ equally-sized steps, denoted as $(z_1, z_2, \dots, z_m)$.
This yields a sequence of prefixes $p_1, p_2, \dots, p_m$, where $p_i =\mathbf{z}[:i]$ is the concatenation of the first $i$ steps.
Similarly, we define a suffix of the gold-CoT $\mathbf{o}^*$ starting at step $t$ as $s_t = \mathbf{o}^*[t:]$.

For each intermediate step $z_i$ (where $i < m$), we compute its reward by evaluating the quality of the full prefix $p_i = (z_1, \dots, z_i)$ that it concludes. This prefix-based evaluation not only assesses $z_i$ within its full contextual history to ensure logical coherence, but also allows the prefix's score to be directly attributed to $z_i$ as the final, decisive step guiding the path forward.

A naive approach would be to measure the conditional probability of generating a gold-CoT suffix given the prefix $p_i$. However, a high probability might arise simply because the suffix itself is a common or high-probability sequence, regardless of the prefix's quality. Following the work of~\cite{xu2025direct}, to isolate the actual contribution of the prefix, we normalize the raw conditional probability by subtracting a baseline. 
This baseline is defined as the probability of generating the same suffix given the initial question $q$ and a masked version of the prefix $p_i$, denoted $p_{\text{masked}}$. 
The resulting score can thus be interpreted as the information gain provided by the final step $z_i$ within the context of its preceding steps.

The reward for step $z_i$, denoted $r_{\text{step}}(z_i)$, is therefore defined by evaluating its corresponding prefix $p_i$ and finding the maximum log-probability gain over all valid suffixes within the definitive gold-CoT $\mathbf{o}^*$:
\begin{equation}
r_{\text{step}}(z_i) := \max_{t} \left( \log \pi_\theta(s_t | q, p_i) - \log \pi_\theta(s_t | q, p_{\text{masked}}) \right)
\label{eq:pfr_step}
\end{equation}

For the final step $z_m$, however, the reward is treated differently.
This step completes the entire reasoning path $z$, and its quality is best assessed by its ability to produce the correct final answer.
For this terminal step, the objective shifts from measuring information gain to ensuring absolute correctness.
Therefore, its reward is defined directly by the conditional log-probability of generating the ground-truth answer $y^*$, given the full reasoning path $z$:
\begin{equation}
r_{\text{step}}(z_m) := \log \pi_\theta(y^* | q, z)
\label{eq:pfr_final}
\end{equation}

\paragraph{Time Complexity Analysis.}
The computational overhead of P2S for a single problem instance is dominated by the number of forward passes ($C_{\text{fwd}}$) through the policy model $\pi_\theta$.
The process involves two main cost components per iteration.
First, the Gold-CoT synthesis requires sampling and filtering $K$ candidate paths, incurring a cost proportional to $K \cdot C_{\text{fwd}}$.
Second, the PFR calculation for a reasoning path with $m$ steps involves a search over suffixes, resulting in a complexity of approximately $O(m^2 \cdot C_{\text{fwd}})$.
Since $m$ is capped by a constant $\text{MAX\_STEP\_NUM}$, this complexity is well-controlled.
Therefore, the total time complexity is $O((K + m^2) \cdot C_{\text{fwd}})$.
This is a manageable trade-off, and the computation is highly parallelizable.


\subsection{Reward Shaping with Step-wise Weighting}
\label{sec:reward_shaping}

A simple averaging of step-wise rewards is suboptimal because it treats all steps equally. Instead, we adopt a strategy that allows the model a ``grace period'' for initial exploration, such as analyzing the problem or self-correcting from early missteps. To implement this, we introduce a weight shaping mechanism that assigns progressively higher importance to later reasoning steps, thereby focusing supervision on the more converged and critical stages of the reasoning process.

To assign greater importance to later reasoning steps, we compute the final sample-level reward, $R_{\text{PFR-w}}$, as a weighted average of the step-wise rewards $r_{\text{step}}(z_i)$. 
The weight for each step, $w_i$, is generated using a monotonically increasing standard sigmoid $\sigma(i)$, ensuring that later steps contribute more significantly to the final reward. The formulation is as follows:
\begin{equation}
R_{\text{PFR-w}} = \frac{\sum_{i=1}^{m} w_i \cdot r_{\text{step}}(z_i)}{\sum_{i=1}^{m} w_i}
\label{eq:pfr_weighted}
\end{equation}


\begin{table*}[htbp]
\small
\setlength{\tabcolsep}{1.35mm}
\begin{tabular}{@{}lcccccccccc@{}}
\toprule
\multicolumn{1}{l|}{\multirow{2}{*}{\textbf{Model}}} & \multicolumn{5}{c|}{\textbf{Drop}}                                                                                                                                                                                                                                                          & \multicolumn{5}{c}{\textbf{MedicalQA}}                                                                                                                                                                                                                                     \\ \cmidrule(l){2-11} 
\multicolumn{1}{l|}{}                                & \textbf{\textbf{ROUGE}} & \textbf{\begin{tabular}[c]{@{}c@{}}\textbf{ACC\textsubscript{Claude}}\end{tabular}} & \textbf{\begin{tabular}[c]{@{}c@{}}\textbf{ACC\textsubscript{GPT}}\end{tabular}} & \textbf{\begin{tabular}[c]{@{}c@{}}\textbf{ACC\textsubscript{Verifier}}\end{tabular}} & \multicolumn{1}{c|}{\textbf{ACC\textsubscript{Avg}}} & \textbf{\textbf{ROUGE}} & \textbf{\begin{tabular}[c]{@{}c@{}}ACC\textsubscript{Claude}\end{tabular}} & \textbf{\begin{tabular}[c]{@{}c@{}}ACC\textsubscript{GPT}\end{tabular}} & \textbf{\begin{tabular}[c]{@{}c@{}}ACC\textsubscript{Verifier}\end{tabular}} & \textbf{ACC\textsubscript{Avg}} \\ \midrule \midrule
\multicolumn{1}{l|}{Qwen2.5-1.5B-Instruct}           
& 42.23& 51.67& 50.75& 49.15 
& \multicolumn{1}{c|}{50.52}
& 40.30 & 19.20 & 19.20 & 27.00 & 21.80

 \\ 
\midrule

\multicolumn{11}{c}{\textbf{\textit{Prompt-Based}}}                                                                                                                           \\ \midrule
\multicolumn{1}{l|}{COT}                             
& 41.97& 45.33& 49.00& 48.85                                    
&\multicolumn{1}{c|}{47.73}
& 40.09 & 20.40 & 21.60 & 26.60 & 22.87

 \\ 
\multicolumn{1}{l|}{Self-Consistency}
& 45.51& 51.17& 52.67& 52.35                                              &\multicolumn{1}{c|}{52.06}
& 38.76 & 14.10 & 17.13 & 23.75 & 18.33
 \\ 
\midrule
\multicolumn{11}{c}{\textbf{\textit{Fine-tuning and RL methods}}}                           \\ \midrule
\multicolumn{1}{l|}{Full-Sft}
& 71.44& 66.00& 64.50& 63.42                                                 
& \multicolumn{1}{c|}{64.64}
& 50.92 & 20.80 & 20.04 & 22.65 & 21.28
                                      \\
\multicolumn{1}{l|}{GRPO}                            
& 70.89  & 60.00  & 62.25  & 62.12                                           & \multicolumn{1}{c|}{61.46} 
&  46.21    &  17.67    &   21.00   &   25.50   & 21.39
                                     \\
\multicolumn{1}{l|}{GRPO+SFT-loss}                   
& 66.18 & 59.50 & 63.00 & 58.90                                               
& \multicolumn{1}{c|}{60.47} 
& 45.79 & 21.00 & 20.40 & 24.15 & 21.85
                                        \\
\multicolumn{1}{l|}{SFT+GRPO}                        
& 75.28  & 66.50  & \underline{70.14}  & \underline{68.55}                    
& \multicolumn{1}{c|}{\underline{68.40}}  
& 50.57 & \underline{23.33} & 20.00 & 23.80 & 22.38
                                    \\

\multicolumn{1}{l|}{General Reasoner}                
& 73.03 & \underline{67.89} & 65.32 & 66.30 & \multicolumn{1}{c|}{66.50}
& \underline{51.57} & 19.18 & 17.20 & \textbf{27.45} & 21.28
\\
\midrule
\multicolumn{11}{c}{ \textbf{\textit{RLPR methods}}}                                             \\ \midrule
\multicolumn{1}{l|}{DRO}                             
& 74.85 & 66.28 & 67.17 & 66.65 & \multicolumn{1}{c|}{66.70}
& 50.52 & 20.11 & 19.20 & 23.50 & 20.94
\\
\multicolumn{1}{l|}{RLPR}                            
& \underline{75.48} & 67.18 & 68.04 & 67.57 & \multicolumn{1}{c|}{67.60}
& 51.14 & 21.16 & 20.75 & \underline{26.92} & \underline{22.94}
\\
\multicolumn{1}{l|}{VeriFree}                        
& 71.98 & 64.42 & 62.17 & 63.40 & \multicolumn{1}{c|}{63.33}
& 51.46 & 21.98 & \underline{21.68} & 22.85 & 22.17
\\
 \midrule

\multicolumn{1}{l|}{\textbf{P2S}}                            
& \textbf{76.78} & \textbf{69.11}  & \textbf{72.14}  & \textbf{70.85}    & \multicolumn{1}{c|}{\textbf{70.70}}
& \textbf{52.90} & \textbf{24.33}  & \textbf{22.67}  & 25.85 & \textbf{24.28}
 \\
\bottomrule
\end{tabular}
\caption{Performance comparison of various Reasoning methods on general-domian QA task. \textbf{Bold} and \underline{underline} indicate the best and second-best results, respectively.}
\label{tab:1}
\end{table*}


\subsection{Hierarchical Reward Integration}

A key advantage of our P2S framework is its flexibility, as the Path Faithfulness Reward ($R_{\text{PFR-w}}$) can function either as a standalone process signal or be integrated with other rewards. 
We present a powerful hierarchical paradigm that combines P2S with an outcome-based reward, assigning scores with a clear priority.
First, malformed trajectories are heavily penalized. 
If any trajectory yields a correct answer, we exclusively use this outcome signal to rapidly amplify the advantage of successful paths. 
Only when all valid paths fail does our dense PFR serve as a fallback, ensuring a fine-grained learning signal is always available to mitigate reward sparsity.

This hierarchical logic can be formalized concisely. Let $F(i) \in \{0, 1\}$ be an indicator function where $F(i)=1$ if the format of trajectory $i$ is correct. Let $S_{\mathcal{G}} = \max_{j \in \mathcal{G}} R_{\text{outcome},j}$ be a binary variable indicating whether any trajectory in the group $\mathcal{G}$ was successful. The final reward $R_i$ for trajectory $i$ is then:

\begin{equation}
R_i = 
\begin{cases} 
-1 & \text{if } F(i) = 0 \\
R_{\text{outcome},i} & \text{if } F(i) = 1 \text{ and } S_{\mathcal{G}} = 1 \\
R_{\text{PFR-w},i} & \text{if } F(i) = 1 \text{ and } S_{\mathcal{G}} = 0
\end{cases}
\end{equation}

\paragraph{Cold-Start.}
To ensure training stability, we adopt a curriculum warm-up strategy~\cite{liu2025ghpo}. For the initial $S_{\text{warmup}}$ training steps, the model learns the basic task structure using only format-based rewards, with our PFR component deactivated. Subsequently, the full P2S reward mechanism is enabled to refine the logical quality of the reasoning process.

\section{Experiments}
\subsection{Experimental Setup}

\subsubsection{Datasets}
We focus on reasoning tasks that lack strict structural verifiers due to their open-ended and stylistically diverse answers, but still possess objectively correct outcomes. Accordingly, we train and evaluate our method on two datasets selected to reflect this challenge. (1) DROP \cite{Dua2019DROP}: A challenging reading comprehension benchmark that requires discrete reasoning over open-domain Wikipedia text, such as arithmetic and sorting. (2) Medical QA \cite{chen2024huatuogpto1medicalcomplexreasoning}: An open-ended medical question-answering dataset derived from challenging medical exams. For both datasets, we  process into a question-answering format and filter to include questions under 2000 and answers between 1-50 characters, creating a 10k/2k random train/test split for each.

\subsubsection{Evaluation Metrics}
Our evaluation employs two complementary metrics for final answers. For lexical similarity, we use \textbf{ROUGE-1 F1} to measure overlap with the ground truth answers. To assess semantic correctness, we use LLM-as-a-Judge~\cite{gu2024survey} to judge semantic equivalence, including: Claude 4 Sonnet (\textbf{ACC\textsubscript{Claude}}), GPT-4o (\textbf{ACC\textsubscript{GPT}}), and a trained 1.5B general-domain Verifier (\textbf{ACC\textsubscript{Verifier}})~\cite{ma2025general}. Finally, we report the mean of these three accuracy scores, \textbf{ACC\textsubscript{Avg}}, as a single, robust measure of correctness.

\subsubsection{Baselines}
We compare our method against several baselines, all built upon the \textbf{Qwen2.5-1.5B-Instruct} model. Full implementation details for all experiments are provided in Appendix A. And our baselines are grouped into three categories. (1) \textbf{Prompt-based methods} that require no fine-tuning: Chain-of-Thought (CoT)~\cite{wei2022chain} and Self-Consistency~\cite{wang2022self}. (2) \textbf{Fine-tuning and RL methods}, including full supervised fine-tuning (Full-SFT) and several GRPO~\cite{shao2024deepseekmath} variants. Standalone \textbf{GRPO}, the two-stage \textbf{SFT+GRPO}, and \textbf{GRPO+SFT-loss} (which integrates off-policy knowledge via an auxiliary SFT loss) all use ROUGE-1 F1 as their outcome-based reward. In contrast, \textbf{General Reasoner}~\cite{ma2025general} also employs GRPO but replaces this reward with judgments from a trained 1.5B LLM verifier that assesses semantic equivalence. (3) \textbf{RLPR-based methods}, which leverage the model's own probabilities for reward, including DRO~\cite{xu2025direct}, the original RLPR~\cite{yu2025rlpr}, and VeriFree~\cite{zhou2025reinforcing}. To ensure a fair comparison and mitigate reward collapse during RL phases, P2S along with the General Reasoner and RLPR-based baselines, adheres to a same cold-start Supervised Fine-Tuning paradigm before RL training~\cite{guo2025deepseek}.


\subsection{Main Results}

Main Results in Table~\ref{tab:1} show our method, P2S, outperforms all baselines on both the DROP and MedicalQA datasets. We can draw several key conclusions from the results:

1) P2S significantly improves general-domain reasoning performance. On DROP, it reaches an ACC\textsubscript{Avg} of 70.70, exceeding the strongest fine-tuned baseline (SFT+GRPO at 68.40) by 2.3 points. This leadership extends to MedicalQA, where P2S achieves an ACC\textsubscript{Avg} of 24.28, outperforming the next best method (RLPR at 22.94) by over 1.3 points.

2) Our core hypothesis—that dense process supervision is critical—is validated by these results. P2S's superiority is particularly clear against RLPR-based methods (e.g., RLPR, VeriFree). On DROP, for instance, P2S surpasses the strongest RLPR-based method (RLPR) by 1.3 points in ROUGE (76.78 vs. 75.48) and by over 3 points in ACC\textsubscript{Avg} (70.70 vs. 67.60). This dual improvement proves that our process-focused supervision not only mitigates the reward sparsity of outcome-only approaches but also guides the model to produce answers superior in both form and substance.

3) P2S outperforms representative fine-tuning and RL paradigms, highlighting the efficacy of verifier-free rewards. On DROP, P2S surpasses all GRPO and RLPR variants. More notably, it outperforms General Reasoner by a significant margin of over 4 points in ACC\textsubscript{Avg} (70.70 vs. 66.50), which uses a 1.5B LLM verifier for its reward signal. This is a crucial finding: our internal, process-based rewards are more effective than guidance from a costly external verifier. Furthermore, the reliability of such verifiers is questionable, as evidenced on MedicalQA. General Reasoner's ACC\textsubscript{Verifier} score (27.45) is substantially inflated compared to judgments from large-scale models like Claude (19.18) and GPT (17.20). This discrepancy underscores the robustness and efficiency of our verifier-free P2S framework, especially in new domains.

\subsection{Ablation Study}
Our ablation study on DROP (Table~\ref{tab:ablation}) validates the contribution of each key component in the P2S framework by systematically removing them from the full model.

Gold-CoT Filtering (GCF) is Crucial. Replacing our Gold-CoT filtering with random path selection (w/o GCF) causes the most substantial performance drop, reducing ACC\textsubscript{Avg} by 4.5 points. This confirms that high-quality, faithful reasoning paths are a critical foundation for effective process supervision.
Path Faithfulness Reward (PFR) is the core contribution. Removing our core PFR component (w/o PFR) results in a 2.3-point decrease in ACC\textsubscript{Avg}. This directly validates the effectiveness of our proposed PFR as a critical component for process supervision.
Advanced Reward Mechanisms are Effective. We also validated our reward design choices. Replacing sigmoid-based weight shaping with simple averaging (w/o RS) drops ACC\textsubscript{Avg} by 2.7 points, confirming the benefit of prioritizing later reasoning steps. Similarly, a naive reward summation (w/o HRI) is less effective than our hierarchical integration, proving the advantage of our dynamic fusion strategy.

\begin{table}[htbp]
\centering
\setlength{\tabcolsep}{0.3mm}
\small
\begin{tabular}{@{}lccccc@{}}
\toprule
\textbf{Model} & \textbf{ROUGE} & \textbf{ACC\textsubscript{Claude}} & \textbf{ACC\textsubscript{GPT-4o}} & \textbf{ACC\textsubscript{Verifier}} & \textbf{ACC\textsubscript{Avg}} \\ \midrule
\textbf{P2S (Full)}      & \textbf{76.78} & \textbf{69.11} & \textbf{72.14} & \textbf{70.85} & \textbf{70.70} \\
\midrule
w/o GCF                  & 71.46          & 64.21          & 67.22          & 67.21          & 66.21          \\
w/o PFR                  & 75.28          & 66.50          & 70.14          & 68.55          & 68.40          \\
w/o RS                   & 74.91          & 64.10          & 69.88          & 69.94          & 67.97          \\
w/o HRI                  & 76.70          & 68.20          & 71.27          & 70.20          & 69.89          \\
\bottomrule
\end{tabular}

\caption{Ablation study of P2S components on DROP. \textbf{P2S (Full)} is our complete model; \textbf{w/o GCF} removes Gold-CoT filtering; \textbf{w/o PFR} removes our core Path Faithfulness Reward; \textbf{w/o RS} removes sigmoid-based reward shaping; and \textbf{w/o HRI} removes hierarchical reward integration.}
\label{tab:ablation}
\end{table}


\subsection{Effect of Model Scale}
To investigate its scalability, we evaluate P2S against the untuned base model and the strong SFT+GRPO baseline at 1.5B and 3B scales on DROP (Table~\ref{tab:size}). Results highlight two key findings. First, P2S shows remarkable efficiency: our 1.5B model (70.70 ACC\textsubscript{Avg}) significantly outperforms the much larger 3B base model (62.77), suggesting our process supervision unlocks capabilities beyond simply scaling parameters. Second, P2S's consistent superiority at both scales confirms it is a robust and effective enhancement across different model sizes.

\begin{table}[h!]
\centering
\small
\setlength{\tabcolsep}{0.25mm}
\begin{tabular}{@{}lcccccc@{}}
\toprule
\multicolumn{1}{c}{\multirow{2}{*}{\textbf{Model Scale}}} & \multicolumn{2}{c}{Base} & \multicolumn{2}{c}{SFT+GRPO} & \multicolumn{2}{c}{\textbf{P2S (Ours)}} \\
\cmidrule(lr){2-3} \cmidrule(lr){4-5} \cmidrule(lr){6-7}
\multicolumn{1}{c}{} & ROUGE & ACC\textsubscript{Avg} & ROUGE & ACC\textsubscript{Avg} & \textbf{ROUGE} & \textbf{ACC\textsubscript{Avg}} \\ \midrule
1.5B & 42.23 & 50.52 & 75.28 & 68.40  & \textbf{76.78} & \textbf{70.70} \\
3B   & 47.96 & 62.77 & 81.16 & 74.41 &  \textbf{82.08} & \textbf{77.30}  \\ \bottomrule
\end{tabular}

\caption{Performance on DROP across model scales.}
\label{tab:size}
\end{table}


\subsection{Analysis on Verifiable Subsets}
We study the effectiveness of P2S on domains with readily available verifiers. To this end, we created two verifiable subsets—DROP-verifiable (5k) and MedicalQA-verifiable (2.35k)—by filtering for instances with single-word answers. On these, we compare P2S against two outcome-only baselines: the probabilistic RLPR and the rule-based RLVR.

\begin{figure}
    \centering
    \includegraphics[width=1\linewidth]{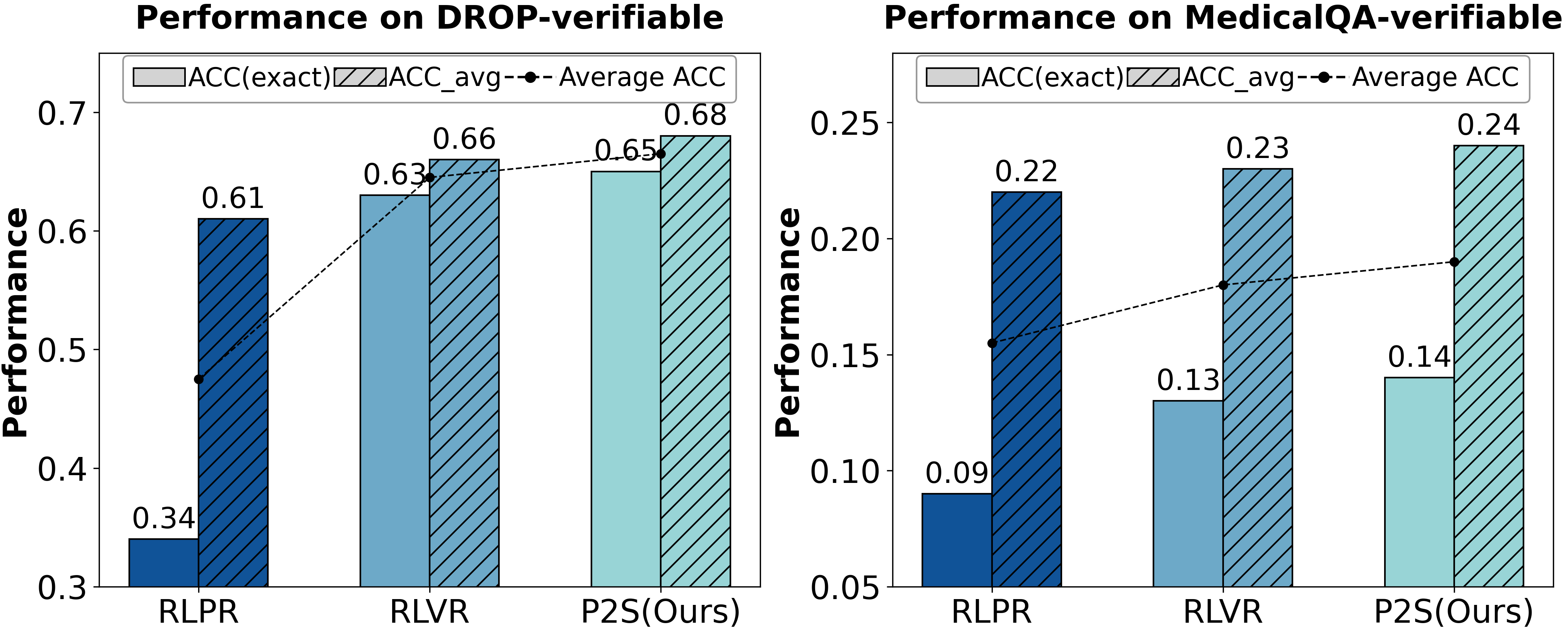}
    \caption{P2S outperforms in verifiable tasks}
    \label{fig:verifiable}
\end{figure}

As shown in Figure~\ref{fig:verifiable}, P2S consistently outperforms both baselines on both subsets, across both exact match (ACC\textsubscript{exact}) and verifier-based average accuracy (ACC\textsubscript{Avg}).This crucial finding proves that our P2S provides a fundamentally superior learning signal, extending its benefits far beyond merely overcoming reward sparsity, even in ideal settings for outcome-only methods.

\subsection{Case Study}
\begin{figure}[h!]
    \centering
    \includegraphics[width=1\linewidth]{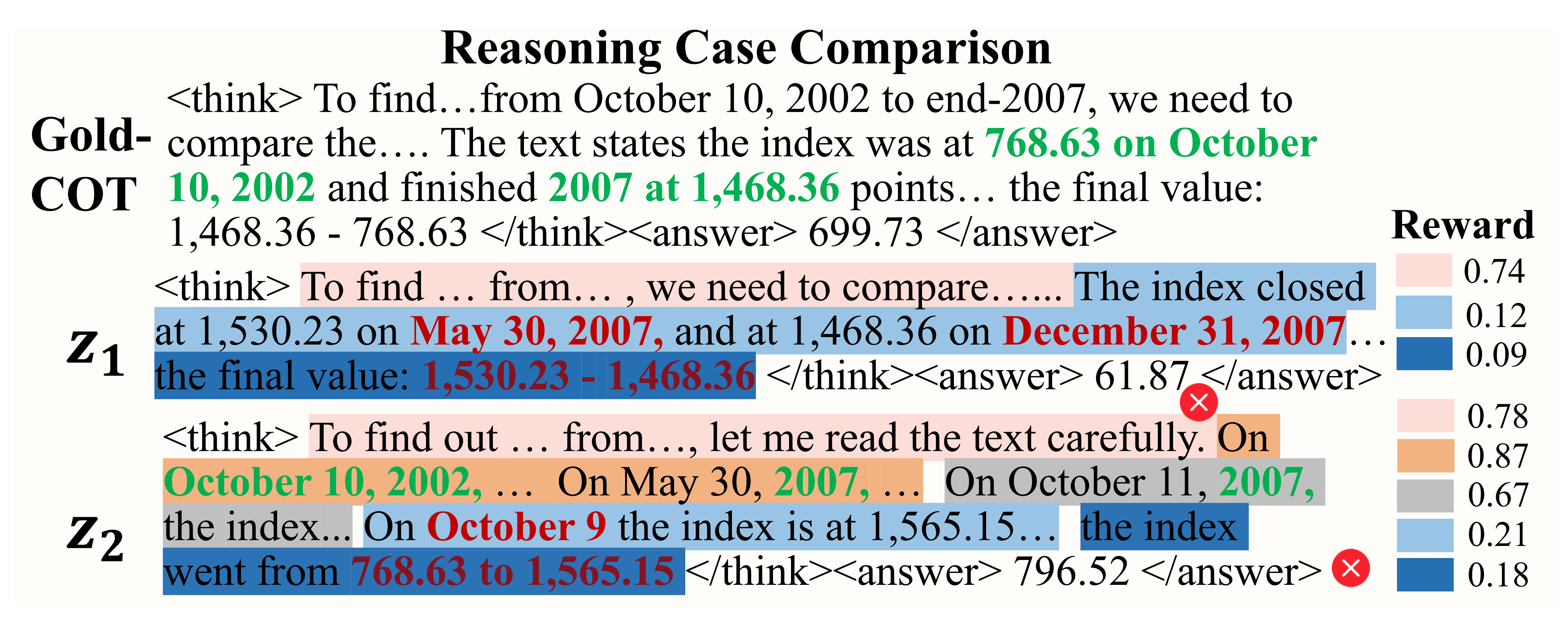}
    \caption{Case Study}
    \label{fig:casestudy}
\end{figure}

Figure~\ref{fig:casestudy} provides a case study to illustrate how our Path Faithfulness Reward (PFR) works. Given a Gold-CoT, we analyze two incorrect reasoning paths, $z_1$ and $z_2$.

In path $z_1$, the model makes an early error by analyzing the wrong dates (highlighted in light blue), leading to a low reward score for that step (e.g., 0.12). The error propagates, resulting in even lower scores for subsequent steps (0.09). In contrast, path $z_2$ correctly identifies the initial entities (highlighted in orange), and our PFR mechanism appropriately assigns a high reward to this correct step (0.87). 

Although both paths ultimately fail to produce the correct final answer, our PFR is capable of discerning valuable, correct sub-steps within an overall incorrect reasoning process. This fine-grained reward allows our framework to reinforce partially correct reasoning even within failed attempts.

\section{Conclusion}
In this paper, we introduced Probabilistic Process Supervision (P2S), a novel, low-cost self-supervision framework. At its core, P2S leverages two key innovations: a dynamic mechanism for synthesizing high-quality Gold-CoTs and the Path Faithfulness Reward (PFR), which provides a dense, step-by-step signal by measuring the faithfulness of a generated reasoning path to a reference. 
Our extensive experiments demonstrated that P2S significantly outperforms strong baselines on challenging reasoning benchmarks. This work proves that it is both feasible and effective to learn directly from the reasoning process itself without external reward models or human annotation.

\section{Acknowledgments}
This work was supported in part by the ``Pioneer'' and ``Leading Goose'' R\&D Program of Zhejiang (2025C02037), the National Natural Science Foundation of China (62376243, 62406287), Key R\&D Program of Hangzhou (2025SZDA0254), and Ant Group, Chongqing Ant Consumer Finance Co. All opinions in this paper are those of the authors and do not necessarily reflect the views of the funding agencies.

\bibliography{aaai2026}

\clearpage 

\appendix

\appendix
\setcounter{secnumdepth}{2}
\section{Experimental Details}

\begin{figure}[h!]
    \centering
    \includegraphics[width=1\linewidth]{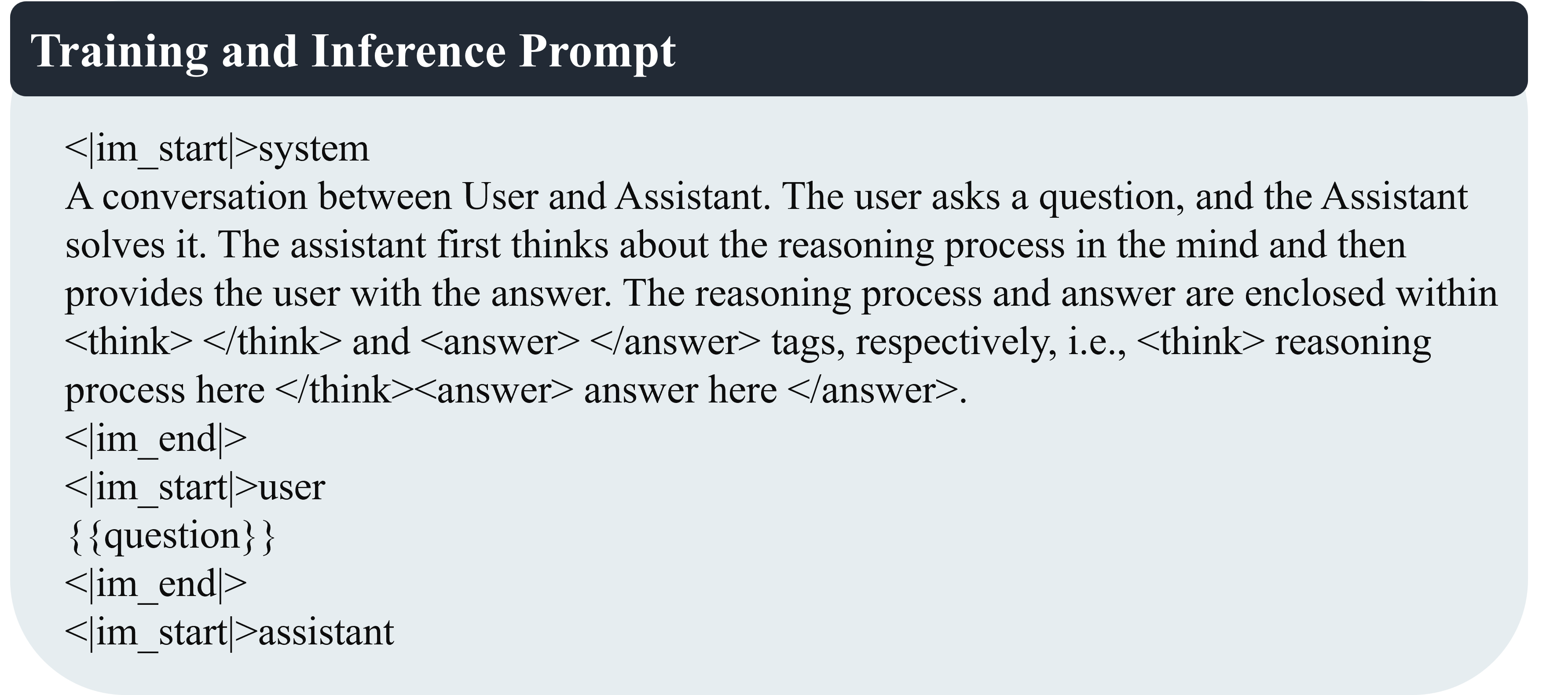}
    \caption{We adopt the training and inference prompt of R1 ~\cite{guo2025deepseek}}
    \label{fig:prompt}
\end{figure}

Our experiments, including P2S and all baselines, are conducted on Qwen2.5-1.5B-Instruct\citep{qwen2025qwen25technicalreport} if not additionally specified. We conducted our experiments using the openr1 ~\cite{openr1} codebase and the TRL ~\cite{vonwerra2022trl} framework. We are thankful for these open-source repositories. For training efficiency, we utilize \texttt{bfloat16} precision and enable FlashAttention-2. P2S and all baselines training was performed on 2 powerful H800 GPUs, each equipped with 80GB of memory and high memory bandwidth. The prompt template is shown in figure~\ref{fig:prompt}.

\subsection{Implementation Details and Hyperparameters}

Key hyperparameters for our main experiment are as follows. We set the learning rate to \texttt{3.0e-6} with a cosine learning rate scheduler and a warmup ratio of 0.1. We use a per-device training batch size of 16 with 8 gradient accumulation steps, resulting in an effective batch size of 256 using a
temperature of 1. To ensure fairness, we maintain 4 samples per prompt for all RL-trained models. Unless otherwise specified, the implementations of all baseline methods follow their original papers and official codebases. For the RL rollout phase, inference is accelerated by vLLM, which is configured to use 80\% of the GPU memory. With
minimal truncation observed, the maximum prompt and completion lengths are set to 1024 and 2048 tokens, respectively. We train 500 steps for all RL models and three epochs for SFT models. For reproducibility, all runs use a fixed random seed of 42. In our method, the $S_{\text{warmup}}$ is configured to 20. For reliable answer extraction, we adopt the “$<$think$>$Reasoning$<$/think$>$$<$answer$>$Answer$<$/answer$>$” template of R1 ~\cite{guo2025deepseek} during training and use the striped content inside answer tags as the generated answer.

\begin{figure}
    \centering
    \includegraphics[width=1\linewidth]{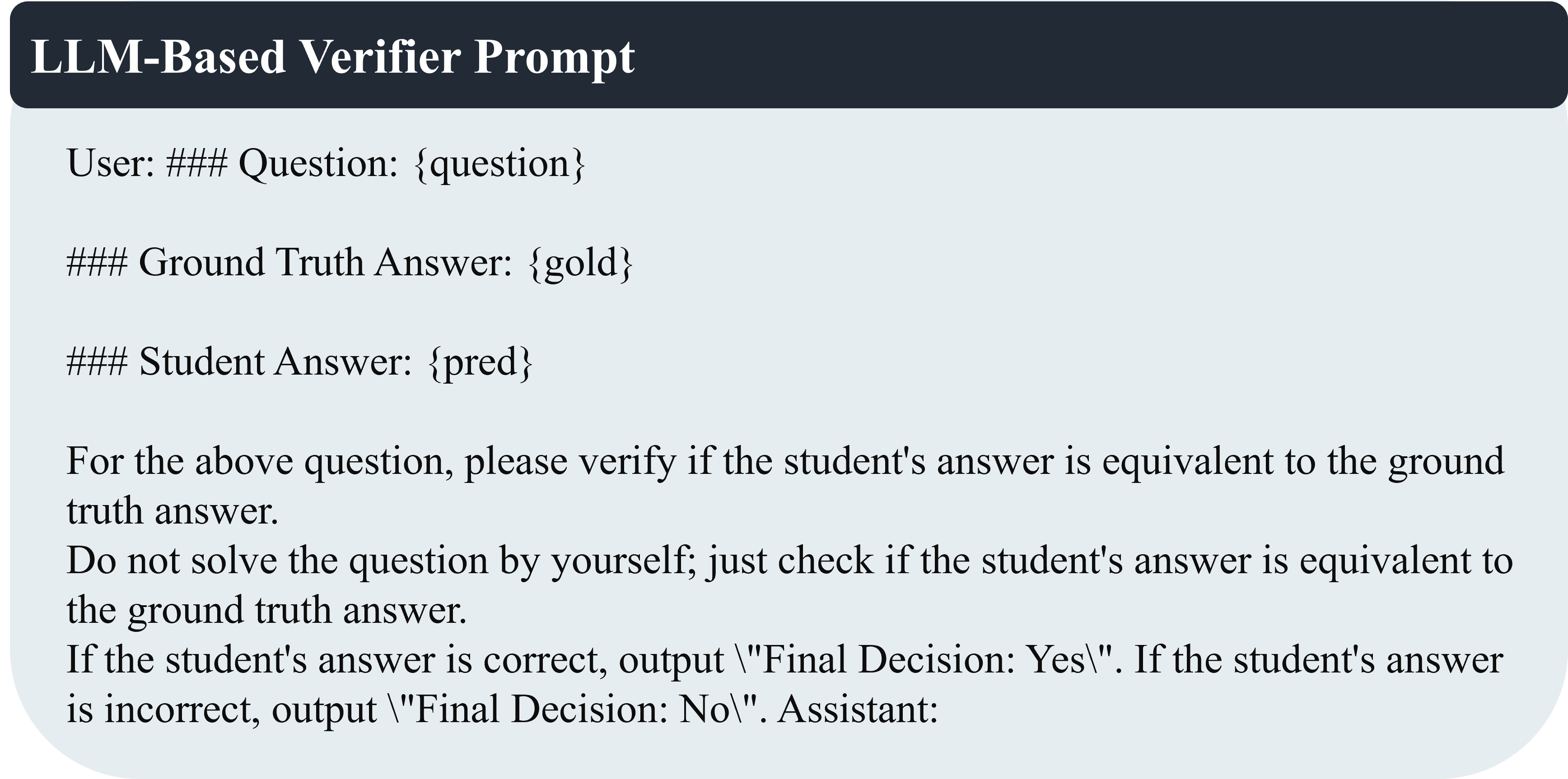}
    \caption{LLM-Based Verifier Prompt}
    \label{fig:verifier_prompt}
\end{figure}

The prompt used for all our large model-based verifiers is detailed in Figure~\ref{fig:verifier_prompt}.

\begin{figure}[h!]
    \centering
    \includegraphics[width=1\linewidth]{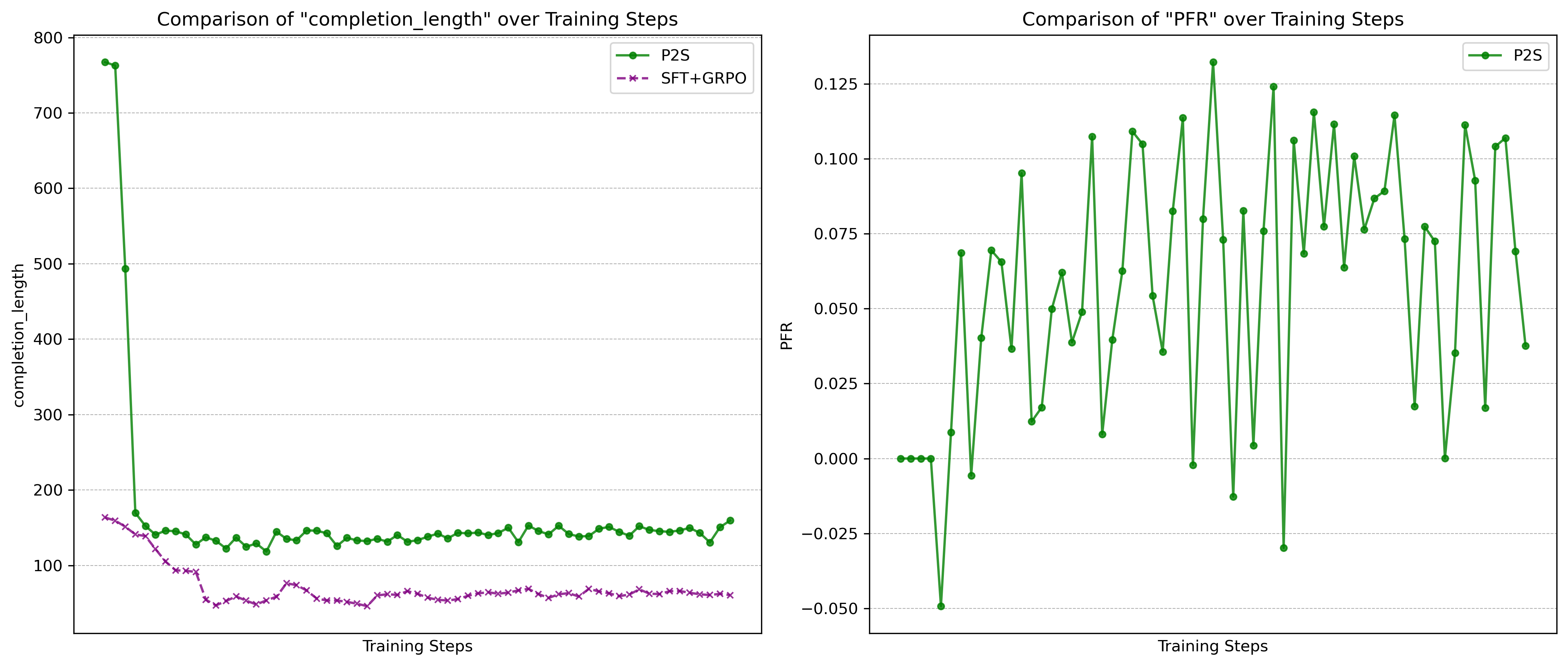}
    \caption{completion step reward comparison}
    \label{fig:completion_step_reward_comparison}
\end{figure}

\begin{figure}[h!]
    \centering
    \includegraphics[width=1\linewidth]{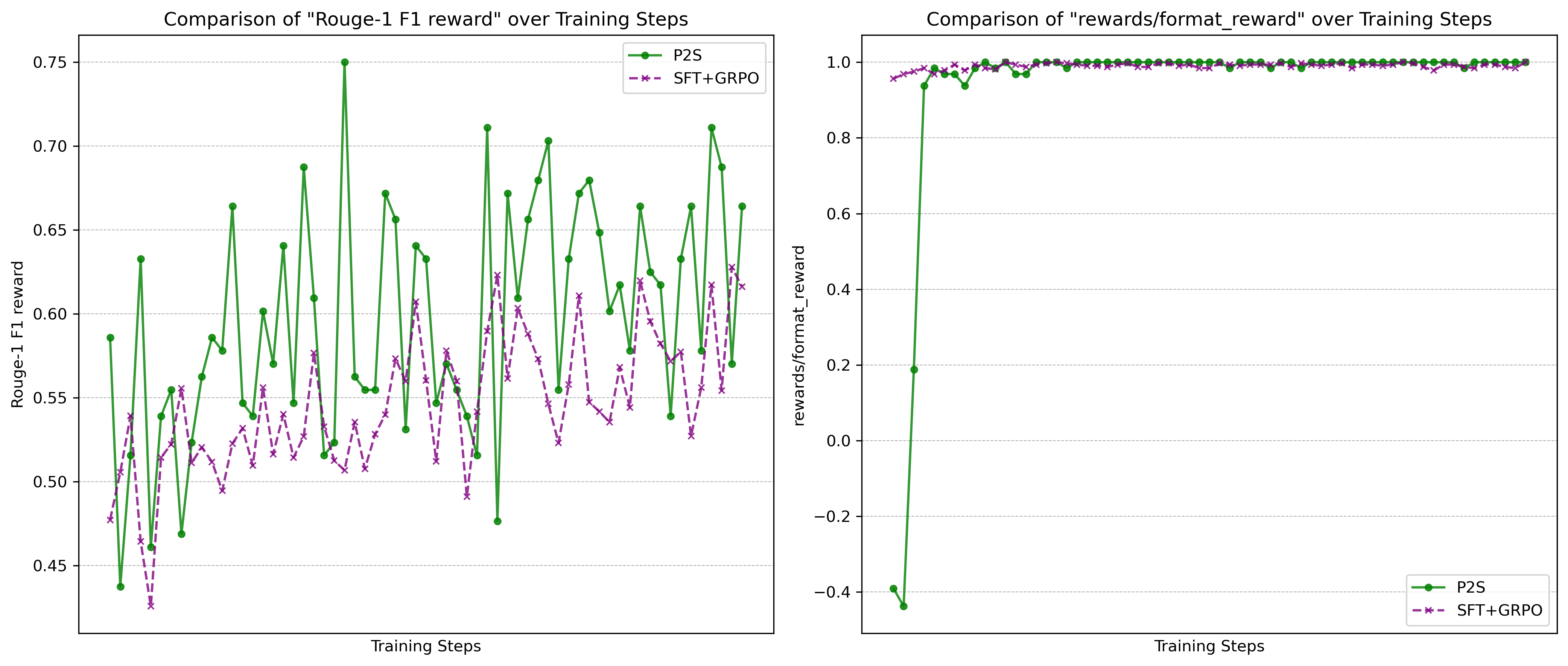}
    \caption{rewards comparison}
    \label{fig:rewards_comparison}
\end{figure}

\subsection{Training Dynamics Analysis}

We analyze the training dynamics in Figure ~\ref{fig:completion_step_reward_comparison} and ~\ref{fig:rewards_comparison}. Figure ~\ref{fig:completion_step_reward_comparison} illustrates the internal characteristics of our P2S method compared to the SFT+GRPO baseline. The completion length (left) of P2S stabilizes at a significantly higher level (approx. 150 vs. 60 tokens), encouraging more detailed reasoning. Concurrently, its unique Path Faithfulness Reward (PFR, right) remains consistently positive, indicating that the model is effectively learning to generate faithful reasoning steps.

These strong internal signals translate to superior external performance, as shown in Figure ~\ref{fig:rewards_comparison}. While both methods quickly master the required response format (right), P2S consistently achieves a higher Rouge-1 F1 reward (left). This demonstrates that our framework's ability to foster longer and more faithful reasoning directly results in higher-quality final outputs

\section{Algorithm Workflow}
The complete workflow is outlined in Algorithm ~\ref{alg:appendix_p2s_complete}.

\begin{algorithm*}[t]
\caption{The P2S Algorithm Flow}
\label{alg:appendix_p2s_complete}
\begin{algorithmic}[1]
    \renewcommand{\algorithmicrequire}{\textbf{Input:}}
    \renewcommand{\algorithmicensure}{\textbf{Output:}}
    \REQUIRE Query $q$, reference answer $y^*$, policy model $\pi_\theta$, number of candidates $K$, current training step $S_{\text{current}}$, the set of rollouts $\mathcal{R}_{z}$ generated for the query $q$ by the policy model $\pi_\theta$ at the step $S_{\text{current}}$.
    \ENSURE The set of rewards $\{R_z\}_{z \in \mathcal{G}}$ for each path in the group. 
    
    \STATE // -------------------- \textbf{Phase 1: Dynamic Gold-CoT Synthesis and Filtering} -------------------- 
    \STATE Generate a set of $K$ candidate paths: $\mathcal{C} \leftarrow \{\mathbf{o}_k \sim \pi_\theta(\cdot | q, y^*)\}_{k=1}^K$.
    \STATE Filter for format-correct paths: $\mathcal{C}_{\text{formatted}} \leftarrow \{\mathbf{o}_k \in \mathcal{C} \mid \text{$R_{format}$}(\mathbf{o}_k) == 1\}$.
    \IF{$\mathcal{C}_{\text{formatted}} \neq \emptyset$}
        \STATE For each candidate in $\mathcal{C}_{\text{formatted}}$, we compute a quality score $S_k = \sum_{t=1}^{|y^*_k|} \log \pi_\theta(y^*_{k,t} | \mathbf{q}, \mathbf{z}_k, y^*_{k,<t})$
        \STATE Select 
        \item[]  $\mathbf{o}^* \leftarrow \underset{\mathbf{o}_k \in \mathcal{C}_{\text{formatted}}}{\arg\max} \, S_k$
    \ELSE
        \STATE $\mathbf{o}^* \leftarrow \text{null}$. \COMMENT{No valid candidates were found.}
    \ENDIF
    \STATE // -------------------- \textbf{Phase 2: Reward Calculation for a Generated Path} --------------------
    \FOR{each path $z$ in the group $\mathcal{G}$} 

        \STATE // -------------------- Hierarchical Reward Logic and Cold Start for current path $z$ --------------------
        \IF{Format of $z$ is invalid}
        
        \STATE $R_z \leftarrow -C_{\text{penalty}}$; \textbf{continue} 
        \ENDIF

        \STATE Let $S_{\mathcal{G}} = 1$ if any path in the batch $\mathcal{G}$ produced the correct answer, else $S_{\mathcal{G}} = 0$.
        \IF{$S_{\mathcal{G}} == 1$}
            \STATE $R_z \leftarrow R_{\text{outcome}}(z)$; \textbf{continue} 
        \ENDIF
        \IF{$S_{\text{current}} < S_{\text{warmup}}$}
            \STATE $R_z \leftarrow 0$; \textbf{continue} 
        \ENDIF
        \IF{$\mathbf{o}^* == \text{null}$}
            \STATE $R_z \leftarrow 0$; \textbf{continue} 
        \ENDIF

        \STATE // --------------------  PFR Calculation for current path $z$ --------------------
        \STATE Segment $z$ into $m$ steps $(z_1, \dots, z_m)$.
        \STATE Let $p_i = (p_1, \dots, p_{m})$ be the prefix of the path.
        \STATE Let $s_t = (s_1, \dots, s_{|o^*|})$ be the suffix of the gold-CoT path $o^*$.
        \FOR{$i = 1$ to $m$}
            \IF{i $<$ m}
            \STATE Compute  $r_{\text{step}}(z_i) := \max_{t} \left( \log \pi_\theta(s_t | q, p_i) - \log \pi_\theta(s_t | q, p_{\text{masked}}) \right)$
            \ELSIF{i == m}
            \STATE Compute  $r_{\text{step}}(z_m) := \log \pi_\theta(y^* | q, z)$
            \ENDIF
        \ENDFOR
        
        \STATE Compute step weights using the sigmoid function: $w_i = \sigma(i) = \frac{1}{1 + e^{-i}}$ (for $i=1, \dots, m$).
        \STATE Compute  $R_{\text{PFR-w}} = \frac{\sum_{i=1}^{m} w_i \cdot r_{\text{step}}(z_i)}{\sum_{i=1}^{m} w_i}$. 
        \STATE $R_z \leftarrow R_{\text{PFR-w}}$.
    \ENDFOR 
\STATE \textbf{return} $\{R_z\}_{z \in \mathcal{G}}$ 
\end{algorithmic}

\end{algorithm*}

\end{document}